\documentclass[journal=jcisd8,manuscript=article]{achemso}

\usepackage{chemformula} 
\usepackage[T1]{fontenc} 

\usepackage{xcolor} 
\usepackage{booktabs}
\usepackage{makecell}
\usepackage{setspace}

\usepackage{hyperref}
\AtBeginDocument{\singlespacing}

\author{Michael L. Parker$^\dagger$}
\email{michael@optibrium.com}
\author{Samar Mahmoud$^\dagger$}
\author{Bailey Montefiore$^\dagger$}
\author{Mario Öeren$^\dagger$}
\author{Himani Tandon$^\dagger$}
\author{Charlotte Wharrick$^\dagger$}
\author{Matthew D. Segall$^\dagger$}
\affiliation[Optibrium]
{$^\dagger$Optibrium Ltd., F10-13 Blenheim House, Cambridge Innovation Park, Denny End Road, Cambridge, CB25 9GL, UK}

\title[Improving Predictions of Molecular Properties with Graph Featurisation]
  {Improving Predictions of Molecular Properties with Graph Featurisation and Heterogeneous Ensemble Models}

\begin{document}







\begin{abstract}

  We explore a "best-of-both" approach to modelling molecular properties by combining learned molecular descriptors from a graph neural network (GNN) with general-purpose descriptors and a mixed ensemble of machine learning (ML) models. We introduce a MetaModel framework to aggregate predictions from a diverse set of leading ML models. We present a featurisation scheme for combining task-specific GNN-derived features with conventional molecular descriptors. 
  
  We demonstrate that our framework outperforms the cutting-edge ChemProp model on all regression datasets tested and 6 of 9 classification datasets. We further show that including the GNN features derived from ChemProp boosts the ensemble model's performance on several datasets where it otherwise would have underperformed. We conclude that to achieve optimal performance across a wide set of problems, it is vital to combine general-purpose descriptors with task-specific learned features and use a diverse set of ML models to make the predictions.
  
\end{abstract}

\section{Introduction}

Machine learning (ML) has been widely adopted in modern drug discovery for various use cases, including property prediction, molecular generation and optimisation, and synthesis prediction\cite{Vamathevan2019, Dara2021}. The methods used range from simple linear regression to huge deep-learning models with hundreds of millions of parameters \cite{Irwin2022}, and have greatly enhanced chemists' ability to work with molecular data, particularly at large scales.

A key challenge in optimising predictive model performance when working with molecular data is choosing the input descriptors (commonly referred to as features in ML) used to train the model\cite{CarracedoReboredo2021}. These descriptors are typically numerical or categorical values corresponding to simple molecular properties, such as molecular weight or number of atoms within a structure, which are likely useful when we want to predict a more complex or less well-understood property. We typically want a descriptor set that contains as much relevant information as possible about the molecules, but if we make the descriptor set too large we invite over-fitting, where the ML model effectively memorises the training data and fails to generalise to new data.

In recent years, graph neural networks (GNNs) have shown great potential in molecular property estimation, notably the PotentialNet and ChemProp packages\cite{potentialnet, chemprop1, chemprop2}. The key advantage of GNNs is their ability to learn relevant structural information about molecules for a given problem. This offers great flexibility when compared to a standard set of fixed descriptors, which are unlikely to be optimised for a specific problem. On the other hand, learned features may be unable to derive key molecular properties because of limitations in the network architecture or training data. This suggests that combining learned, task-specific GNN features with general-purpose descriptors will likely outperform either individually.

GNN implementations typically combine the graph network with a Feed-Forward Network (FFN) head to convert their latent molecular representations into actual predictions; this is effectively the same as a simple multi-layer perceptron, representing a potential limiting factor for their performance.
Despite the great success demonstrated by artificial neural networks (ANNs) in many areas, they still generally underperform at the apparently simple task of predicting properties from tabular data because they have an inductive bias towards overly smooth functions, and unlike tree-based models, struggle to ignore uninformative features\cite{Grinsztajn22}. 

In general, no single class of models will be optimal on every dataset, or even on every endpoint within the same dataset, given the range of scales and other properties of datasets available in practical use-cases (referred to as the no-free-lunch theorem\cite{Wolpert2002}). 
All ML models have their own inductive biases and are therefore better suited to different problems.  For example, random forest models will always produce output functions with a series of discrete steps, while neural networks will prioritise smoother functions. This could, for example, mean that ANNs struggle to model sharp transitions like activity cliffs, while random forests struggle to interpolate smoothly between sparse data points. Additionally, if all models are members of the same class, these biases will extend to estimates of uncertainty calculated from model ensembles. 
We hypothesise that while GNNs are capable of deriving task-specific information that is unlikely to be present in general-purpose descriptor sets and, therefore, give competitive performance, they are likely unable to make optimal use of standard fixed descriptors on some datasets, giving them a significant disadvantage over other models, such as gradient boosting or Gaussian processes. To ensure the best possible performance on the broadest possible range of tasks, it is therefore essential to include multiple different model classes in the modelling process.

One potential method to achieve this is to use a mixed ensemble of ML models, combining their predictions into a final (ideally more accurate and less biased) estimate. There are no fixed rules for constructing ensemble models, so various approaches have been explored in the literature\cite{Rifkat21}. Many authors have reported slight improvements in model performance by fitting an ensemble of the same model, but the best performance is typically found by aggregating diverse models. For example, \citet{Kwon2019} used a sophisticated ensemble method using an ANN to aggregate predictions from a selection of different predictive algorithms, trained on different descriptor sets, and to predict different targets. The aggregation model then learned to extract the most meaningful information from all predictions for each target. \citet{Vo2023} took a similar approach, using an ANN to combine various ML model predictions with similarity data to predict drug-drug interactions effectively. 


In this paper, we aim to explore the potential for leveraging learned representations from GNNs with a featurising and modelling framework that allows a wide range of ML algorithms to be applied to a task and benefit from learned molecular features. We hypothesise that classical ML algorithms will outperform FFNs on many tasks, particularly when they have access to learned molecular representations. Specifically, we want to answer the following questions:
\begin{itemize}
    \item Does a mixed model ensemble with access to task-specific GNN-derived features and general-purpose descriptors outperform one without GNN features?
    \item Can such a model outperform the GNN-based model used to derive the features, when both have access to the same information?
    \item Can we find an effective hyperparameter optimisation strategy that can be used to select the GNN parameters to optimise for the performance of the final ensemble model?
\end{itemize}

\section{Ensemble modelling}
\label{sec:metamodel}

Ensemble learning is common in machine learning and is frequently used with ensembles of weak learners (such as decision trees) that make up a strong learner (e.g., a random forest or gradient boosting model). The predictive performance of the ensemble is typically significantly better than that of the individual learners, and estimates of the model's confidence can be made from the level of agreement between the weak learners.

Uniform ensembles of strong learners with different initialisations can also offer modest improvement over a single strong learner. However, these models will still have the same inductive biases and thus will tend to fit the same functional forms and make the same errors.
We can take this approach one step further and build a non-uniform ensemble of different strong learners (for example, combining ANNs and tree-based models) with their outputs weighted by their performance on validation data. Ideally, this would ensure that, for any given problem, we have several models in the ensemble with the optimal inductive biases and near-optimal hyperparameters, without having to run hyperparameter optimisation. A secondary benefit of this is that it provides a trivial method to estimate the model uncertainty from the variance in sub-model predictions, and can take model bias into account in the process (this is outside the scope of the current work, but is a promising avenue for future research).

We construct a meta-model (hereafter referred to as `the MetaModel') from a set of classical ML models (referred to as sub-models hereafter, full details in the supplementary information). To increase the independence of the sub-model predictions, we give each sub-model a different randomised train/validation split. The contributions of the sub-model predictions are weighted by the inverse of the mean-squared error (MSE) on the validation set for regression models and by either the area under the curve (AUC) of the receiver operating characteristic (ROC) or precision-recall curve (PRC) for classification models, with PRC being used when the classes are heavily skewed (set at when the minority class has <10\% of the data points). 

The sub-models are described in detail in the supplementary information. In brief, for the regression ensemble, we use Lasso regression, ridge regression, K-nearest neighbours (KNN), kernel-ridge regression (KRR), random forests (RF), gradient boosting with XGBoost, Gaussian processes (GP), radial basis function interpolation (RBF), multi-layer perceptrons (MLP), and ResNet models. For classification, we use KNN, linear and quadratic discriminant analysis (LDA/QDA), logistic regression (LogReg), na\"{i}ve Bayes (NB), RF, XGBoost, GP, MLP, and ResNet models.

When fitting, we train all sub-models in the ensemble on their individual training sets, evaluate their performance on their respective validation sets, and prune the ensemble down to the ten best-performing sub-models. We then prune the feature set down to just those with $>2\%$ the importance of the most important feature, and retrain the smaller ensemble on the same training data. The MetaModel performance is then evaluated on the test set. Sub-model predictions are weighted based on their validation set performance. We combine numerical predictions as a weighted mean of the sub-model predictions, and combine categorical predictions as weighted means of the probabilities returned by the sub-models.  
When calculating feature importances for the model, we use the weighted mean of the normalised sub-model feature importances, calculated either with their in-built methods or by using permutation importance (derived using the SHAP library\cite{shap}).

\section{Featurising}

\subsection{External descriptors}

We initially featurise the molecules using the \texttt{CalcMolDescriptors} function in RDKit\cite{rdkit}, dropping any descriptors that return NaN or Inf values for a given dataset, and scale the descriptor values to unit variance and zero mean. This is not strictly optimal as some descriptors are not normally distributed (see discussion in \citet{chemprop1}); however, as we will be comparing the performance of two different modelling approaches on the same set of descriptors, we only care about their relative performance, and optimal descriptor scaling is not necessary. We manually exclude two descriptors, Ipc and Kappa3, which can return extreme values on specific molecules. Ipc returns extremely large ($>10^{30}$) values for large molecules, and Kappa3 can return large values ($>9000$) for very small molecules (it divides by the number of length 3 paths in the molecule plus the Hall-Keir $\alpha$\cite{Hall91}, so this term can be very small if there are no such paths). 

\subsection{Learned descriptors}

\begin{figure*}
    \centering
    \includegraphics[width=\linewidth]{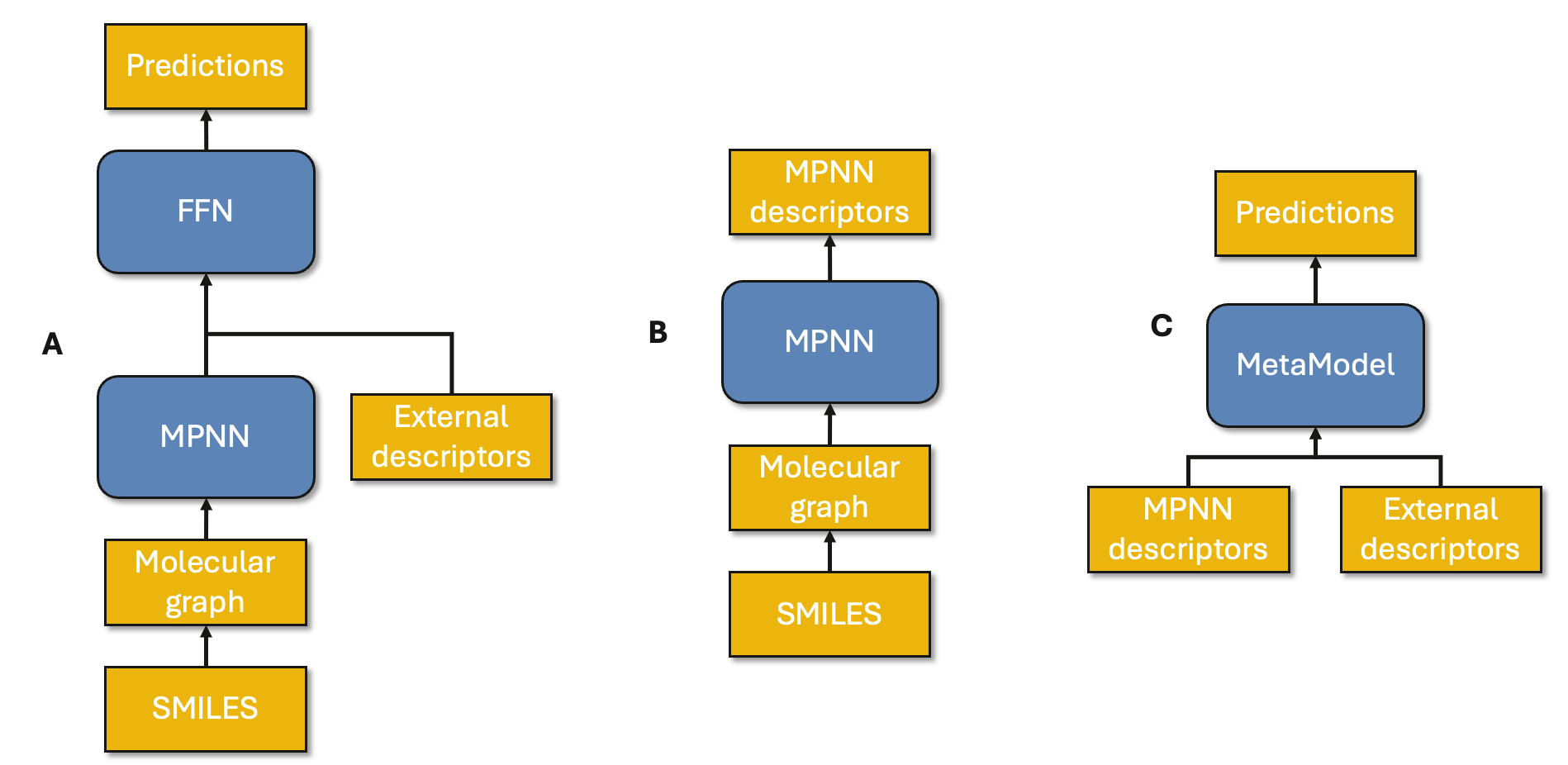}
    \caption{The process used to featurise molecules using ChemProp. A) A ChemProp model, consisting of an MPNN that derives a latent molecular representation and an FFN head that makes predictions, is trained to predict molecular properties from a dataset of molecules and (optional) external descriptors. B) We calculate a new set of features (referred to as MPNN descriptors) by taking the output of the MPNN with the FFN head removed, after training in step A. C) These new descriptors are combined with (optional) external descriptors and used to train a MetaModel comprising diverse ML sub-models.}
    \label{fig:featurisation}
\end{figure*}

We use a message-passing neural network (MPNN)\cite{Gilmer17} from the ChemProp library\cite{chemprop2} version 2.1.2 as a second featurisation step. ChemProp is commonly used for molecular property prediction and is available as an efficient Python package, which makes it ideal for integration into our featurisation pipeline. The basic outline of the process for using a ChemProp MPNN as a featuriser is shown in Fig.~\ref{fig:featurisation}. 

Because of regularisation in the training process, some of the learned features produced by the MPNN are zero for all rows in the training data. These should be trivial to filter out in general, and in the MetaModel case they are caught by the feature pruning step. See the Discussion on Task Specific Descriptors for more detail.

\section{Results}

\subsection{Datasets}
\label{sec:datasets}
We evaluate the model performance on the benchmarking datasets from MoleculeNet \cite{moleculenet}, excluding the QM7b dataset (which only contains 3D coordinates without SMILES strings) and the PDBBind dataset (which includes both protein and ligand structures, and is outside the scope of this work). These datasets cover various target variables (referred to as tasks), including classification and regression problems, over a wide range of data scales. Some datasets contain only a single task, whereas others have multiple targets to be derived. We summarise the datasets in the supplementary information.

\subsection{Baseline results}
\label{sec:baseline}

First, we perform a baseline evaluation of the models on every dataset. We train ChemProp with the default hyperparameters and with external descriptors, and evaluate its performance on the test set. We then extract the latent representation and combine it with the external descriptors to train the MetaModel. When training ChemProp, we use checkpointing to save and retrieve the model from the best-performing training epoch (evaluated on the validation set), then use this model iteration for testing. In each case, we use the recommended data splitting strategy from MoleculeNet\cite{moleculenet} to divide the data into train/test/validation sets. Note that while the splitting methods are the same as those used in MoleculeNet, the exact splits differ because we used a custom pipeline to process and featurise the data rather than using DeepChem. This is reasonable because we are primarily interested in the relative performance of the MetaModel and ChemProp on the same test set. We are not comparing with models derived elsewhere. For the MetaModel, each sub-model is randomly assigned its own train/validation split using the same splitting strategy. In every case, we evaluate both models' performance on the same test set using the metric recommended by MoleculeNet\cite{moleculenet}.

\begin{figure}
    \centering
    \includegraphics[width=\linewidth]{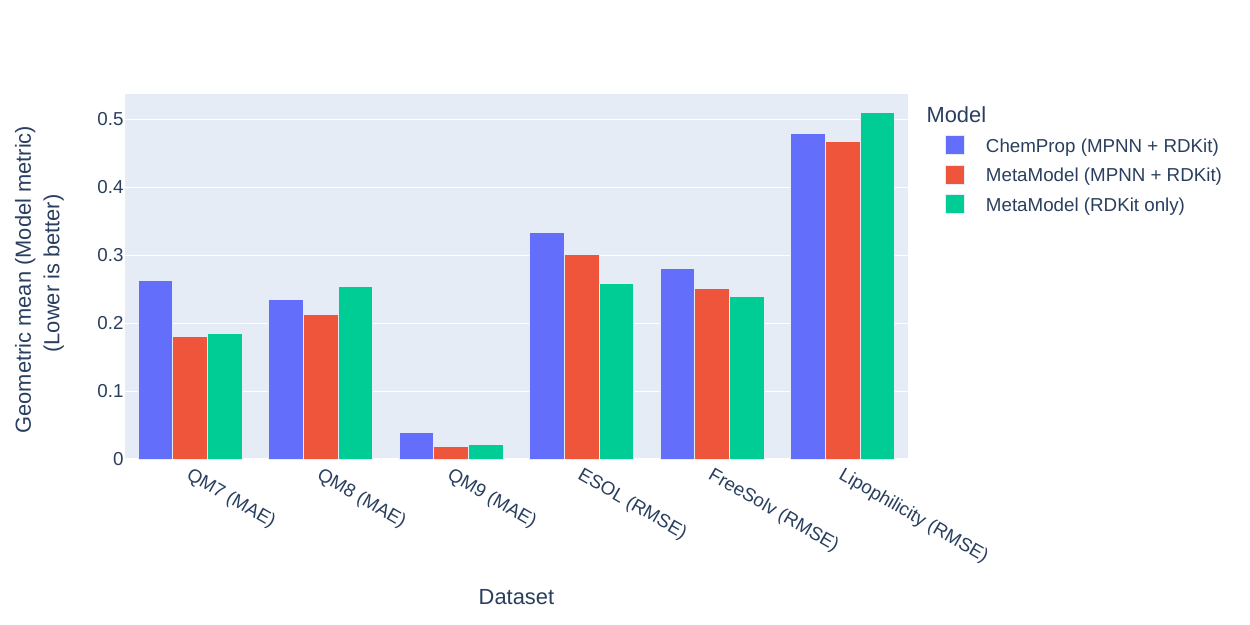}
    \includegraphics[width=\linewidth]{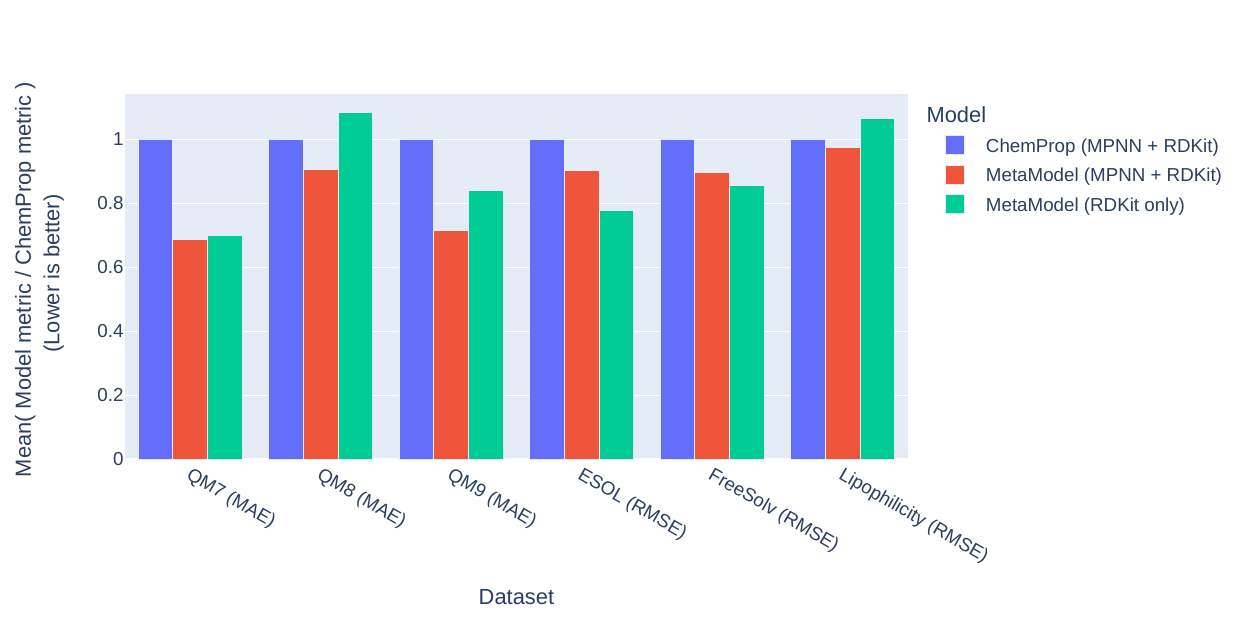}
    \caption{Mean relative performance of ChemProp and the MetaModel (with and without ChemProp features) for each regression dataset. The metric used in each case (as recommended in MoleculeNet) is shown in brackets next to the dataset name. In every case, lower is better. Where multiple targets are present in a dataset, we first calculate the relative metrics, then the geometric mean. We use the geometric mean as some datasets contain MAE/RMSE values for different targets that differ by several orders of magnitude. The top panel shows the absolute values, and the bottom panel shows them normalised to the ChemProp mean.}
    \label{fig:reg_baseline}
\end{figure}

\begin{figure}
    \centering
    \includegraphics[width=\linewidth]{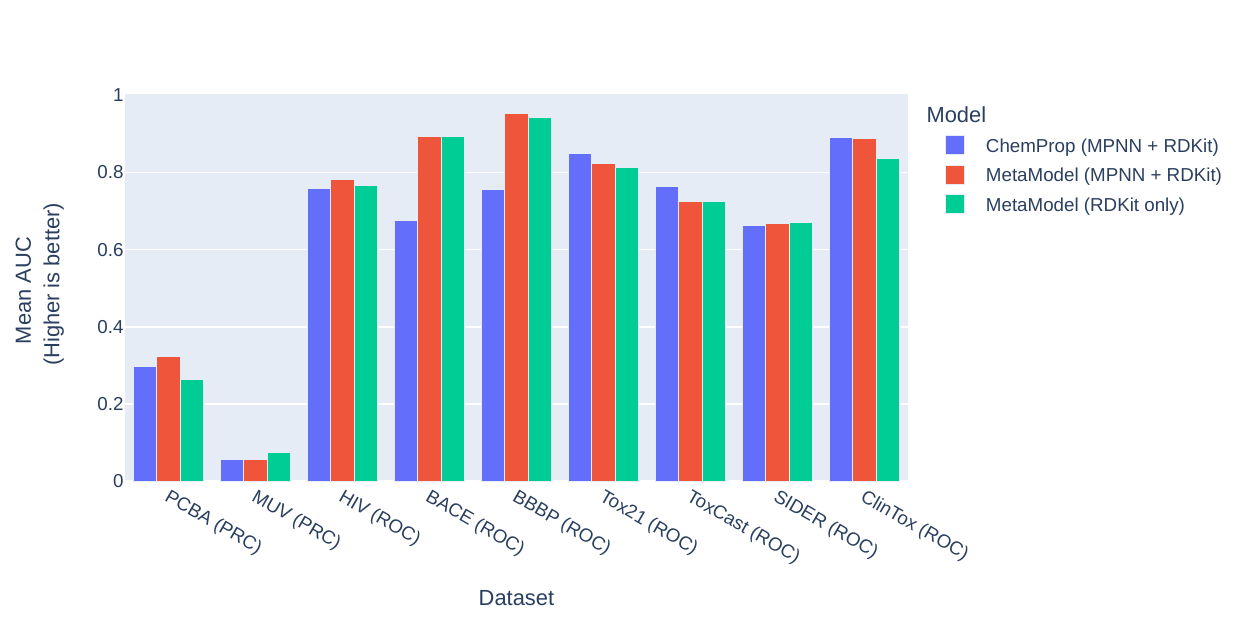}
    \includegraphics[width=\linewidth]{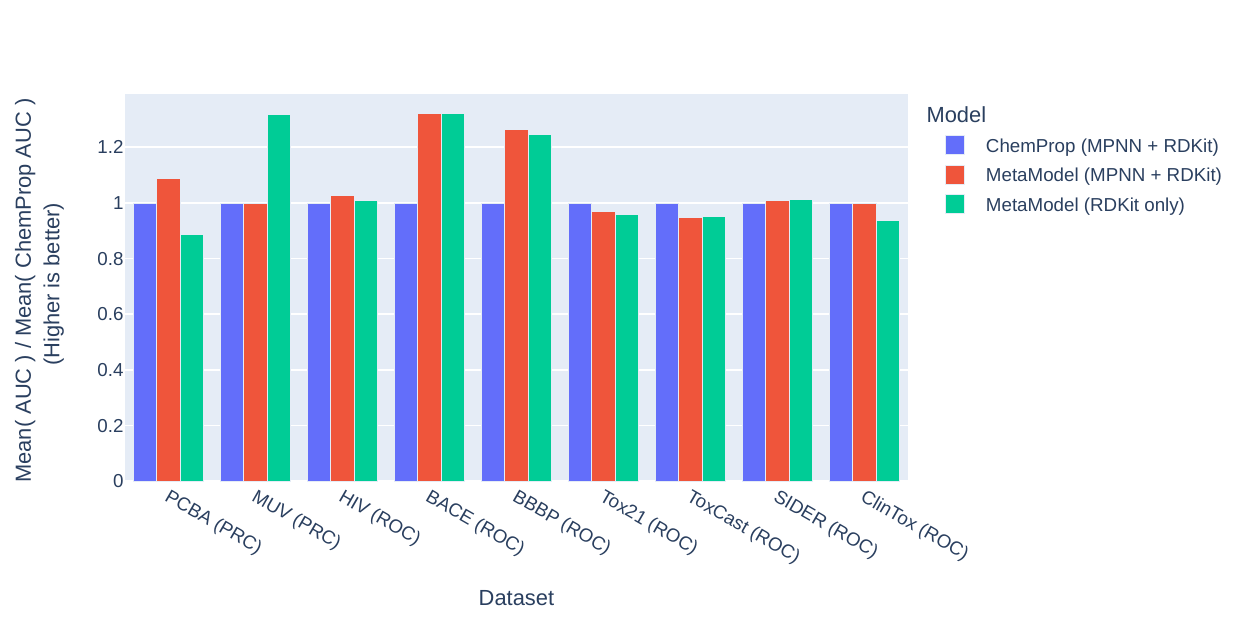}
    \caption{Model performance on the classification datasets, averaged over all targets for each dataset. The top panel shows the absolute values, and the bottom panel shows values normalised by the ChemProp AUC. In every case, metrics are shown in brackets, and higher is better. Where multiple targets are present, we calculate the arithmetic mean AUC value for each model separately, then calculate the ratios of the means. This avoids division by zero or very small AUC values skewing the result.}
    \label{fig:class_baseline}
\end{figure}

The results for regression datasets are shown in Fig.~\ref{fig:reg_baseline}, classification datasets in Fig.~\ref{fig:class_baseline}, and all results are summarised in Table~\ref{tab:baseline_results}. In the regression case, where we report an average metric for a dataset with multiple target variables, we use the geometric mean. This is because the MAE and MSE can differ by orders of magnitude within a single dataset (most notably QM9), and taking the arithmetic mean neglects this. This is not necessary for the classification datasets, as the AUC is a linearly distributed statistic between 0 and 1 (the difference between an MAE of 0.01 and an MAE of 0.001 is important; the difference between an AUC of 0.01 and an AUC of 0.001 likely is not).

In the regression case, the MetaModel outperforms ChemProp on all 6 datasets, with high significance (evaluated by bootstrapping the model predictions, see supplementary information) on 5 and marginal significance $(p=0.05)$ on 1. The MetaModel with ChemProp and RDKit input features outperforms the MetaModel with just RDKit features significantly on 4 of the 6 datasets, and outperforms it with low significance ($p=0.21$ and $p=0.18$) on 2 more. On two datasets, adding ChemProp descriptors does not improve the fit, and in one case it leads to a significantly worse result ($p=0.01$). In three datasets, adding the ChemProp descriptors significantly improves the MetaModel fit.

The MetaModel outperforms ChemProp on 6 out of 9 on the classification datasets, although only 3 of these are significant. ChemProp significantly outperforms the MetaModel on two datasets, ToxCast and Tox21. Both datasets are sparse, with many gaps and multiple target variables (617 and 12, respectively). We suggest that this advantages a multi-target model like ChemProp, which can learn common patterns across different targets and from every row of the dataset, over a single target model that can only learn from a subset of the data. We explore this further in the discussion. The MetaModel with access to ChemProp features outperforms the MetaModel without them on 6 of 9 datasets, although only 3 of these are significant at the 90\% level (P=0.00, 0.02, and 0.09 for PCBA, ClinTox and Tox21, respectively). In no case does the MetaModel without ChemProp descriptors significantly outperform the MetaModel with them.

\begin{table}
\caption{Summary of baseline model performance on MoleculeNet datasets. For the regression datasets, we report the geometric mean of the relevant error metric (lower is better), and for the classification datasets we report the mean AUC (higher is better). We report the results for each dataset for both ChemProp and the MetaModel, with and without ChemProp descriptors. In all cases, the better-performing model is highlighted in bold. Note that the MAE and RMSE are calculated after the data are standardised to a zero mean and unit standard deviation. The $p$ values 1, 2 and 3 correspond to the probability that the MetaModel (RDKit+CP) performs worse than ChemProp, the probability that the MetaModel (RDKit) performs worse than ChemProp, and the probability that the MetaModel (RDKit+CP) performs worse than the MetaModel (RDKit). Probabilities are calculated using bootstrapping on the predicted test set values (see Supplementary Information).}
\label{tab:baseline_results}
\begin{tabular}{lcccccc}
\toprule
Dataset & \makecell{MetaModel \\ (RDKit+CP)} & \makecell{MetaModel \\ (RDKit)} & ChemProp & $p_1$ & $p_2$ & $p_3$ \\
\midrule
QM7 (MAE) &  \textbf{0.181}  &  0.185  &  0.264 & 0.0 & 0.0 & 0.28 \\
QM8 (MAE) &  \textbf{0.213}  &  0.255  &  0.235 & 0.0 & 1.0 & 0.0\\
QM9 (MAE) &  \textbf{0.019}  &  0.022  &  0.040 & 0.0 & 0.0 & $2.4\times10^{-15}$ \\
ESOL (RMSE) &  0.302  & \textbf{ 0.259}  &  0.334 & 0.18 & 0.01 & 0.99 \\
FreeSolv (RMSE) &  0.252  &  \textbf{0.240}  &  0.280 &0.21 & 0.05 & 0.85 \\
Lipophil. (RMSE) &  \textbf{0.467}  &  0.511  &  0.479 &$1.3\times10^{-4}$ & 0.93 & $2.1\times10^{-5}$\\
\midrule
PCBA (PRC) &  \textbf{0.326}  &  0.273  &  0.298  & $8.6\times10^{-8}$ & 1.0 & 0.0\\
MUV (PRC) &  0.056  &  \textbf{0.074}  &  0.056 & 0.41 & 0.34 & 0.55\\
HIV (ROC) &  \textbf{0.781}  &  0.766  &  0.760  & 0.21 & 0.38 & 0.21\\
BACE (ROC) &  \textbf{0.893}  &  0.892  &  0.675  & $2\times10^{-4}$ & $3\times10^{-5}$ & 0.51\\
BBBP (ROC) &  \textbf{0.954}  &  0.942  &  0.755  & 0.001 & 0.002 & 0.25\\
Tox21 (ROC) &  0.824  &  0.813  &  \textbf{0.849} & 0.99 & 1.0 & 0.09\\
ToxCast (ROC) &  0.725  &  0.726  &  \textbf{0.763} & 1.0 & 1.0 & 0.25\\
SIDER (ROC) &  0.669  &  \textbf{0.671}  &  0.662 & 0.19 & 0.22 & 0.56 \\
ClinTox (ROC) &  0.889  &  0.835  &  \textbf{0.891} & 0.22 & 0.95 & 0.02\\
\bottomrule
\end{tabular}
\end{table}

From our baseline models, we can conclude that giving the MetaModel access to ChemProp's latent representation improves its performance, and that the MetaModel with external descriptors plus ChemProp's latent representation outperforms ChemProp with the same information on most datasets.

We present an analysis of the performance of the different sub-models in the supplementary information.

\subsection{MPNN features only}
\label{sec:mpnn_only}

So far, we have considered the predictions made by ChemProp and the MetaModel based on a combination of learned molecular representations and external descriptors. However, it is interesting to investigate their performance when external descriptors are not included to see how this affects the relative performance of the two models. By seeing which model performs better with which information available, we can infer how well the two models are using the general-purpose and learned features.

We train the models as in the previous section, but exclude the RDKit featurisation step, forcing both models to rely purely on ChemProp's learned molecular representations. For this section, and the other follow-on studies below, we prioritise smaller datasets to manage the computational feasibility of experimentation (as the MetaModel is designed for single-target problems, datasets with $N$ target variables require $N$ models to be trained). In this case, we exclude the QM9, ClinTox, and PCBA datasets.

\begin{figure}
    \centering
    \includegraphics[width=0.8\linewidth]{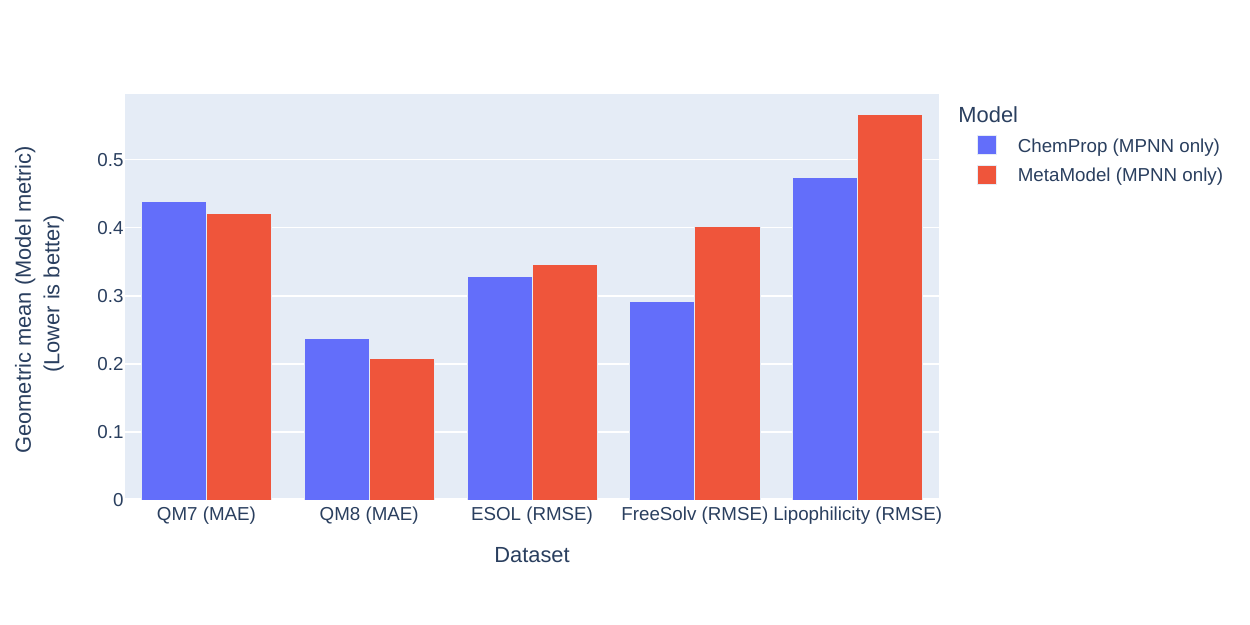}
    \includegraphics[width=0.8\linewidth]{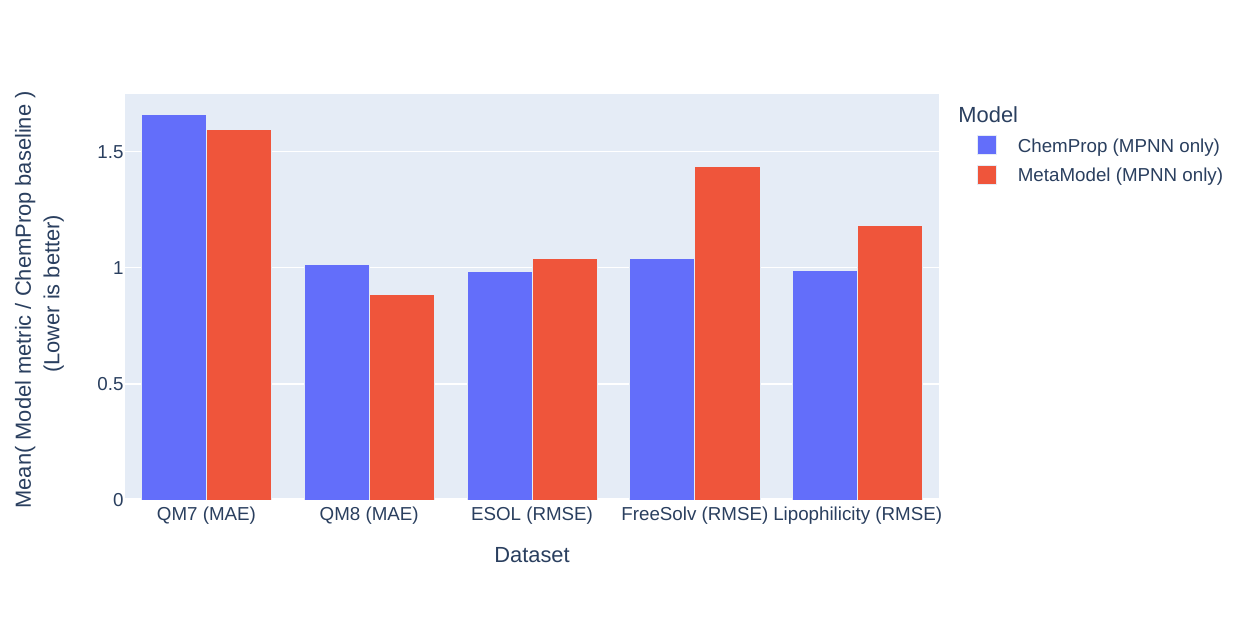}
    \caption{Performance of the MetaModel and ChemProp on the regression datasets with no external descriptors (lower is better). The top panel shows the absolute values, and the bottom panel shows the values normalised to those of the baseline ChemProp model (MPNN + RDKit). Metrics are shown in brackets, and aggregated as in Fig.~\ref{fig:reg_baseline}.}
    \label{fig:regression_noext}
\end{figure}

\begin{figure}
    \centering
    \includegraphics[width=0.9\linewidth]{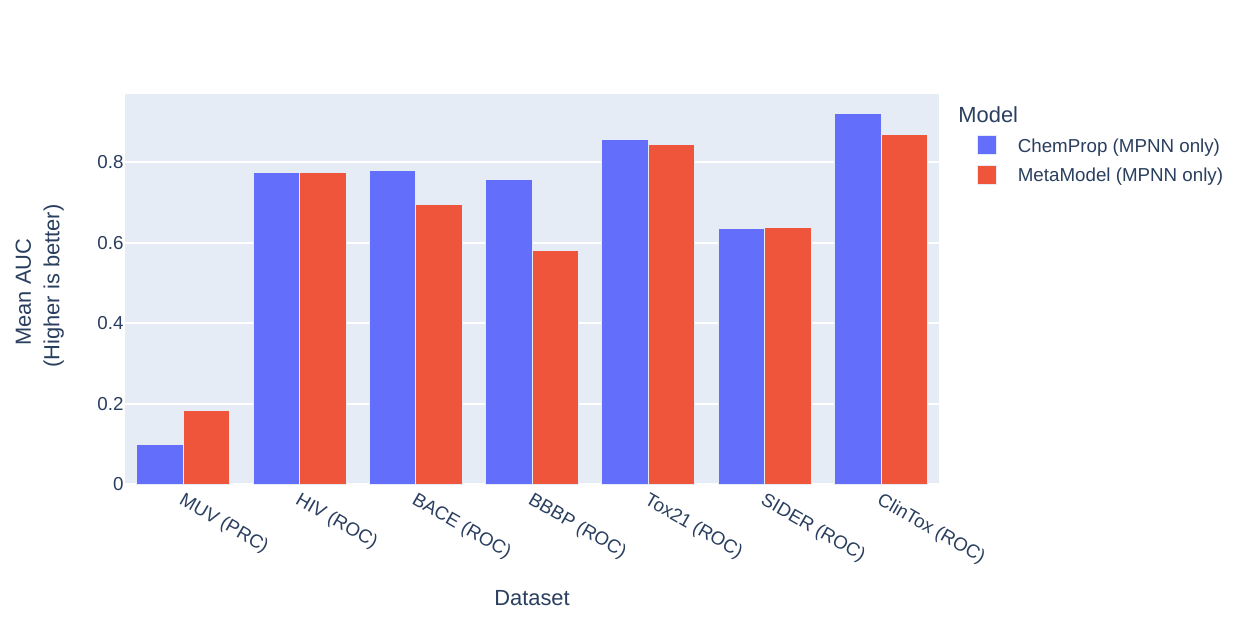}
    \includegraphics[width=0.9\linewidth]{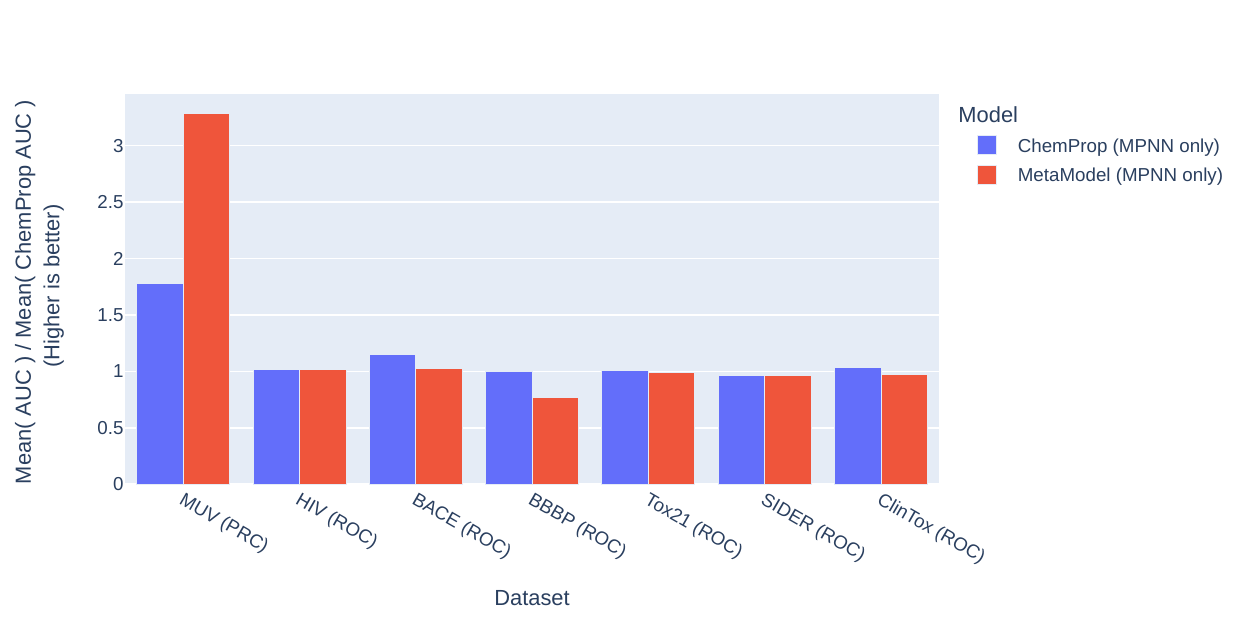}
    \caption{MetaModel and ChemProp performance on the classification datasets with no external descriptors (higher is better, scores are normalised relative to those of the baseline ChemProp model (MPNN + RDKit) in the bottom panel). Metrics are shown in brackets, and aggregated as in Fig.~\ref{fig:class_baseline}.}
    \label{fig:classification_noext}
\end{figure}

The results for this experiment are shown in Figs.~\ref{fig:regression_noext} and \ref{fig:classification_noext} for regression and classification datasets, respectively, and the values are given in Table~\ref{tab:noext_results}. As we might expect, ChemProp now outperforms the MetaModel on most datasets, suggesting that the MetaModel relies more strongly on the external descriptors. In fact, ChemProp does not show a performance drop from removing the RDKit descriptors on most datasets. A notable exception is the QM7 dataset, where both ChemProp and the MetaModel perform dramatically worse (MAEs of 1.66 and 1.60 times the respective baselines). This implies that there are descriptors included in the RDKit features that are both crucial for predicting this dataset and not possible for ChemProp to derive itself.

\begin{table}[h]
\caption{Summary of the model performance when external descriptors are not included. Values in brackets show the model performance relative to the ChemProp baseline model. In each row, the best-performing model is highlighted in bold. $P_1$ and $P_2$ are the probabilities of achieving a statistic as good or better than the observed metric with the corresponding MPNN + RDKit model (i.e. this is the probability that removing the RDKit descriptors has led to no improvement).}
\label{tab:noext_results}
\begin{tabular}{lcccc}
\toprule
Dataset & \makecell{MetaModel} & $P_\mathrm{1}$ & ChemProp & $P_2$  \\
\midrule
QM7 (MAE) &  \textbf{0.42} (1.60) & 1.0 & 0.44 (1.66) & 1.0\\
QM8 (MAE) &  \textbf{0.21} (0.89) & $3.5\times10^{-5}$ & 0.24 (1.01) & 0.96\\
ESOL (RMSE) & 0.35 (1.04) & 0.95 & \textbf{0.33} (0.99) & 0.48 \\
FreeSolv (RMSE) & 0.40 (1.44)& 0.54 & \textbf{0.29} (1.04)& 0.63\\
Lipophil. (RMSE) & 0.57 (1.18)& 0.53 & \textbf{0.47} (0.99)& 0.35\\
\midrule
MUV (PRC) &  \textbf{0.18} (3.29) & $1.5\times10^{-8}$ & 0.10 (1.78) & 0.02 \\
HIV (ROC) &  0.773 (1.02) & 0.62 & \textbf{0.774} (1.02) & 0.29 \\
BACE (ROC) &  0.69 (1.03) & 1.0 & \textbf{0.78} (1.05) & 0.02\\
BBBP (ROC) &  0.58 (0.77)& 1.0 & \textbf{0.76} (1.00)& 0.50\\
Tox21 (ROC) &  0.85 (1.00)& 0.14 & \textbf{0.86} (1.01)& 0.13\\
SIDER (ROC) &  \textbf{0.638} (0.97)& 0.83 & 0.636 (0.96)& 0.95\\
ClinTox (ROC) &  0.87 (0.98)& 0.53 & \textbf{0.92} (1.03)& 0.05\\
\bottomrule
\end{tabular}
\end{table}

The QM8 dataset is also interesting, as the MetaModel improves significantly over the baseline model while ChemProp gets slightly worse, with marginal significance. The change in ChemProp performance is very small ($\sim1\%$), so this can reasonably be attributed to random variance. This improvement in MetaModel performance implies that the RDKit descriptors are unimportant for predicting these targets, and this is backed up by the poor performance of the MetaModel without ChemProp features in the baseline study.

Interestingly, both ChemProp and the MetaModel improve on the MUV dataset, with the MetaModel improving by a larger amount and outperforming ChemProp, despite relying on ChemProp's latent representation. We note that the absolute improvement is relatively small ($\Delta\mathrm{AUC}\sim0.1$), and the relative change is large because the baseline AUC values were so low. It is not apparent why the improvement occurs here, but one possible explanation is that ChemProp was overfitting when it had access to external descriptors, and therefore derived a less informative latent representation, harming the performance of both models. When the external descriptors are removed, the scope for over-fitting is reduced, and the performance of both models improves. If the inductive bias of the MetaModel's submodels is better suited to the highly skewed data than the FFN of ChemProp, this would then allow the MetaModel to make a better prediction from ChemProp's latent representation than ChemProp itself can.

Overall, removing the external descriptors improves Chemprop's performance relative to the MetaModel, but otherwise no clear pattern is apparent. Both models see predictions improve, worsen, and remain unchanged on different datasets. We note that the ChemProp team found a consistent improvement in the model's performance from including external descriptors when averaged across multiple cross-validation folds\cite{chemprop1}, so it is likely that the effective sample size here is too small to see this effect without cross-validation. However, from the relative drop in performance of the MetaModel compared to ChemProp, we conclude that the MetaModel relies more on these descriptors than ChemProp does, and that including them is necessary for the MetaModel performance to be optimal on certain datasets.

\subsection{Hyperparameter optimisation}

So far, we have looked at out-of-the-box performance for both the MetaModel and ChemProp. However, we note that because the MetaModel starts with an ensemble of sub-models and prunes and weights them according to their performance, this could be regarded as a crude form of hyperparameter optimisation. It is also worth noting that any optimisations that benefit the MPNN, resulting in a more informative latent representation, should, in principle, benefit both the MetaModel and ChemProp and are unlikely to make a significant net difference to their performance. 

ChemProp has already demonstrated excellent out-of-the-box performance, and found that, on most of the MoleculeNet datasets, the improvement from hyperparameter optimisation was small (2--5\%)\cite{chemprop1}, which would not make a difference to our conclusions in most cases. However, we are still interested in examining the effect of hyperparameter optimisation, particularly on the performance of the MetaModel trained on the MPNN features. Because of the prohibitive computational costs, we restrict this analysis to the datasets with fewer than 10000 rows and fewer than 100 target variables.

We optimise for the number of layers of the MPNN and FFN, the hidden dimension of the MPNN and FFN, and the dropout fraction of the FFN. We use the \texttt{Optuna} library\cite{optuna} with the tree-structure Parzen estimator\cite{watanabe2023} search algorithm to explore the parameter space, and search for 30 iterations on each dataset. 

Our primary aim in this section is to test different strategies for optimising ChemProp hyperparameters such that the final MetaModel trained on the ChemProp features will perform optimally.
We explore two approaches: first, we optimise ChemProp in isolation without testing the effect on the MetaModel performance. Second, we optimise ChemProp for the performance of the MetaModel trained on the resulting ChemProp features (without optimising any of the MetaModel parameters). To lower the compute costs of training so many models for many test iterations, we run a smaller MetaModel ensemble during the optimisation process. Once the optimal hyperparameters have been selected, we re-run the evaluation with a full MetaModel ensemble. 

As in the baseline study, we include RDKit descriptors for all fits and use the same train/validation/test split for each dataset, ensuring comparable results.

\begin{table}[h]
    \centering
    \caption{Hyperparameter optimisation results. We compare the results for ChemProp and the MetaModel on each dataset for three different trials: no optimisation (the MPNN + RDKit baseline models), optimising ChemProp (CP) performance in isolation, and optimising the performance of a small MetaModel ensemble trained on ChemProp features (CP+MM). In each case, the best result is highlighted in bold (note that in the case of Tox21, no model outperforms the ChemProp baseline). All values are given relative to the baseline ChemProp result. Numbers in brackets are the probability of the respective baseline model returning a metric as good as or better than the optimised model, calculated by bootstrapping the predicted values (see supporting information).}
    \label{tab:hyperpar_results}
    \begin{tabular}{lrrrrr}
\toprule
 & \multicolumn{2}{c}{ChemProp}  & \multicolumn{3}{c}{MetaModel}  \\
    \cmidrule(lr){2-3}                  
    \cmidrule(lr){4-6}
Dataset &  (CP) & (CP + MM) & (None) & (CP) & (CP + MM) \\

\midrule
QM7 & 0.97 (0.04) & 1.38 (1.0) & \textbf{0.69} & 0.78 (1.0) & 0.73 (0.98) \\
ESOL & 1.0 (0.43) & 0.99 (0.25) & 0.91 & 0.88 (0.26) & \textbf{0.82} (0.002) \\
FreeSolv & 0.95 (0.24) & 0.87 (0.03) & 0.85 & 0.71 (0.01) & \textbf{0.81} (0.37) \\
Lipophilicity & 0.97 (0.21) & 1.06 (0.96) & 0.92 & \textbf{0.91} (0.53) & 0.96 (0.95) \\
\midrule
BACE (ROC) & 1.25 (0.001) & 1.17 (0.001)& \textbf{1.32} & 1.30 (0.66)& 1.19 (1.0) \\
BBBP (ROC) & 1.06 (0.24) & 0.74 (1.0)& \textbf{1.26} & 1.12 (0.99) & 0.96 (1.0)\\
ClinTox (ROC) & 1.02 (0.07)& 1.01 (0.26) & 1.00 & 0.99 (0.41) & \textbf{1.05} (0.02)\\
SIDER (ROC) & 0.99 (0.67)& 1.00 (0.61) & 1.01 & \textbf{1.03} (0.07)& 1.01 (0.46)\\
Tox21 (ROC) & 0.98 (0.99)& 0.92 (1.0) & 0.97 & 0.99 (0.10)& 0.97 (0.45)\\
\bottomrule
\end{tabular}
\end{table}

\begin{figure}
    \centering
    \includegraphics[width=0.8\linewidth]{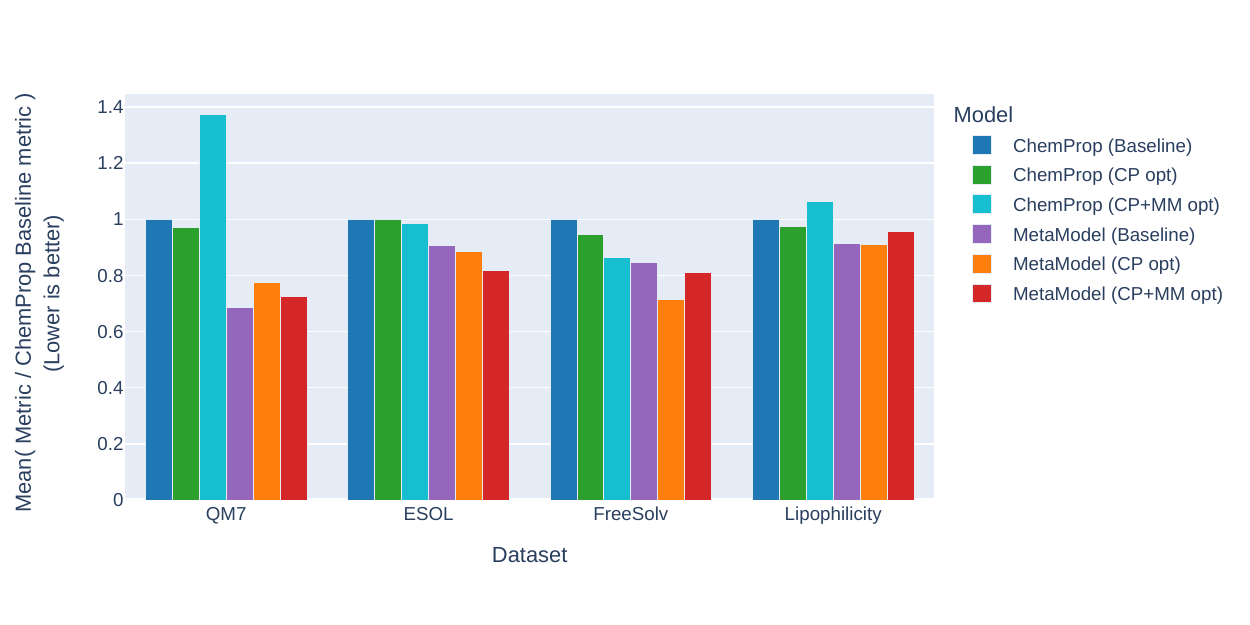}
    \caption{Relative performance of ChemProp and MetaModel with different hyperparameter optimisation schemes on the regression datasets (lower is better). Performance is normalised relative to the out-of-the-box (OOB) ChemProp performance.}
    \label{fig:hyper_reg}
\end{figure}

\begin{figure}
    \centering
    \includegraphics[width=0.8\linewidth]{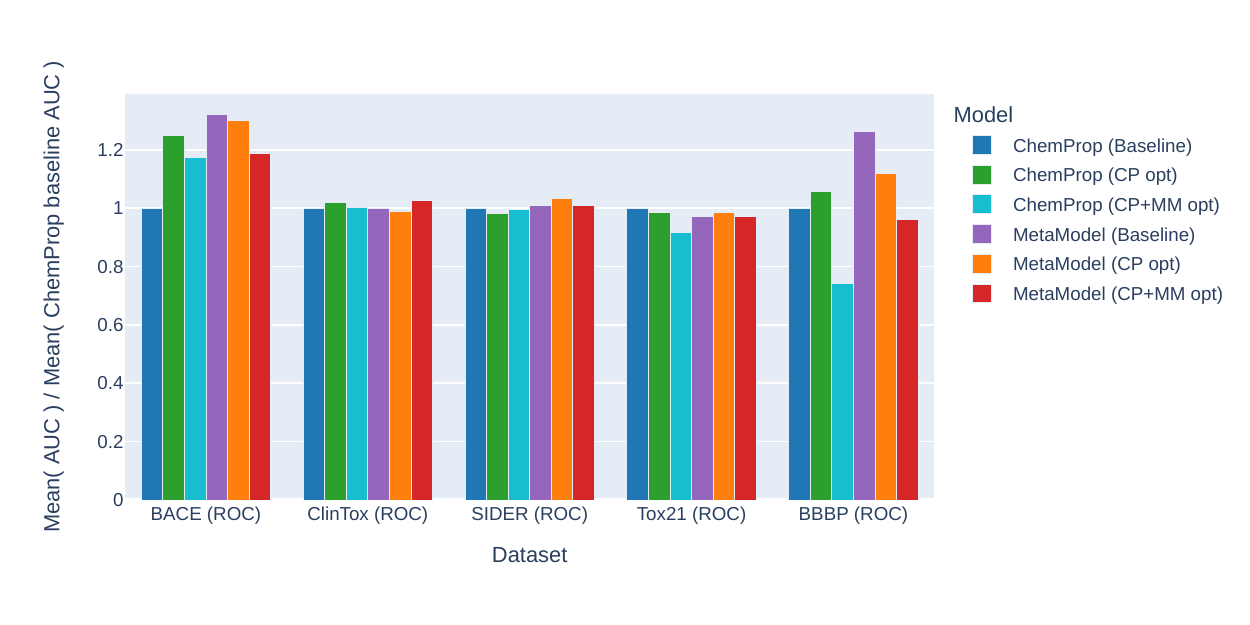}
    \caption{Relative performance of ChemProp and MetaModel with different hyperparameter optimisation schemes on the classification datasets (higher is better). Performance is normalised relative to the out-of-the-box (OOB) ChemProp performance.}
    \label{fig:hyper_cla}
\end{figure}

The results of these optimisation strategies are shown in Figs~\ref{fig:hyper_reg} and \ref{fig:hyper_cla}, and the metrics are given in Table~\ref{tab:hyperpar_results}. No optimisation strategy provides a consistent improvement to the performance of the MetaModel, with the first strategy (optimising ChemProp in isolation) outperforming the baseline on 4 of 9 datasets and the second (optimising ChemProp with the MetaModel) outperforming the baseline on 3 of 9. For ChemProp, the first strategy outperforms the baseline on 6 of 9 datasets and the second on 4. Given the limited size of the datasets, we consider it likely that the potential benefit of optimisation is small compared to the variance in the hyperparameter optimisation process. 

Given the computational cost of running extensive hyperparameter optimisation and the risk of accidentally worsening final performance, we conclude that the best strategy for now is to use the out-of-the-box ChemProp parameters, which reliably perform well across a diverse selection of datasets.

\section{Discussion}

\subsection{Task-specific descriptors}
\label{sec:discuss_learneddescriptors}

The results of our baseline study show that in most, but not all, datasets, adding a learned molecular representation from ChemProp improves the MetaModel fit. 
It seems likely that it is impossible to come up with a general-purpose descriptor set that contains sufficient detail to give optimal performance on any dataset, while also not providing large amounts of irrelevant information that leads to overfitting and poor generalisation. GNN features, derived from a regularised, task-specific neural network, are almost guaranteed to focus only on relevant molecular features, resulting in a more focused descriptor set. 

It may be possible to match or exceed the performance gain from including GNN features through careful manual selection and curation of predefined descriptors; however, this would be very time-consuming and highly skilled work, requiring an expert computational chemist to select, derive, transform and evaluate the features for every target variable in every dataset.
We therefore conclude that learned molecular representations are a vital part of state-of-the-art predictive modelling, particularly at scale. 

\begin{figure}[h]
    \centering
    \includegraphics[width=\linewidth]{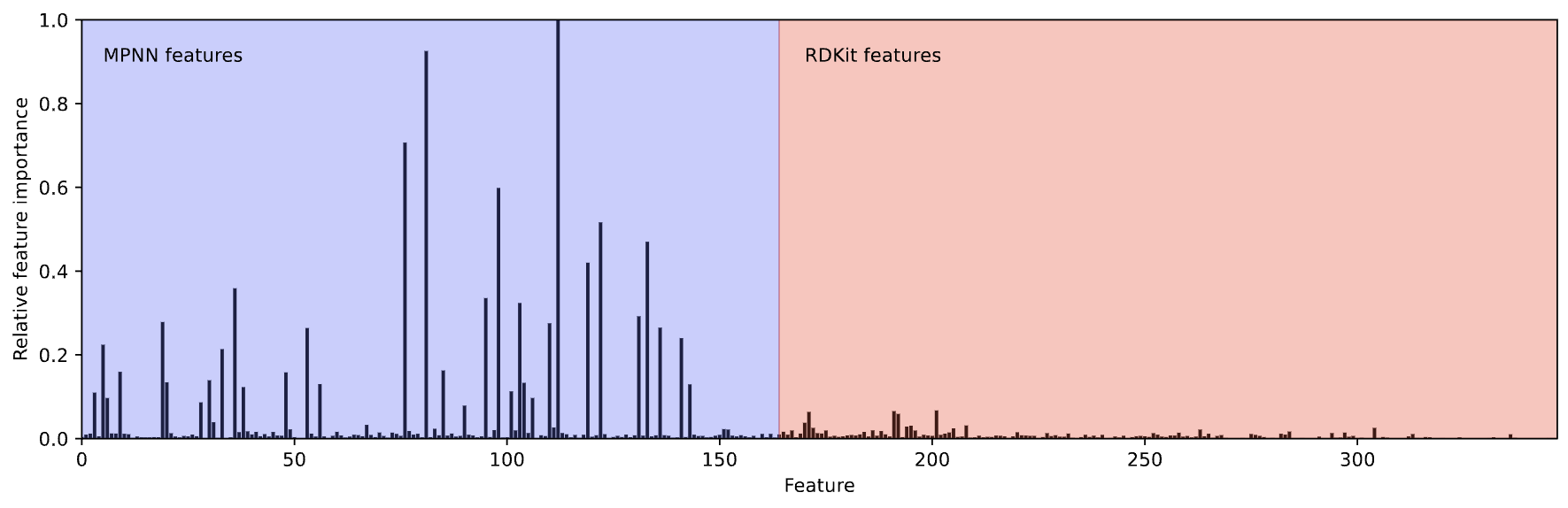}
    \caption{Feature importance of descriptors for MetaModel predictions of the BACE dataset. We exclude all features that are zero for every row in the training set.}
    \label{fig:feature_importance}
\end{figure}

 Further evidence supporting the value of learned features comes from the models themselves.
 Figure~\ref{fig:feature_importance} shows the feature importance for the MetaModel trained on the BACE dataset, calculated from a weighted sum of the normalised feature importances of the individual sub-models in the final ensemble. It is immediately apparent that the model is weighting the MPNN features much more highly than the RDKit features. We speculate that, because the MPNN features are derived with this specific problem in mind, they correlate more simply with the target variable. In contrast, the pre-defined RDKit descriptors are more general and are likely to be used to refine the initial estimates derived primarily from MPNN features. Interestingly, in this case the final model performance is almost exactly the same when only RDKit descriptors are used (AUC of 0.892, compared to 0.893 with both). This implies that there is at least some redundancy between the two feature sets, and that while the MPNN features may be more 'intuitive' for a model to learn, they are not necessarily required.
 
 We also note that nearly half of the MPNN features were zero for all rows in the training set. This is presumably due to regularisation in the ChemProp training process, which aims to prevent overfitting by pushing model weights towards zero. This is not necessarily a problem when using these features, but it does mean care should be taken to exclude uninformative features. 

A secondary question is why general-purpose descriptors are still necessary to achieve the best performance, if learned molecular representations are so powerful and dominate the feature importance. One possibility is that specific descriptors are difficult or impossible for the MPNN architecture to derive. For example, GNNs tend to focus on local structural information in molecules, and can struggle to propagate signals over the full size of the molecule to derive global properties. Similarly, some molecular properties may be too complex to derive from a small dataset on the fly. In this case, having access to these properties without having to derive them is a clear advantage. We suggest that this is the cause of the dramatic increase in error observed on the QM7 dataset for both ChemProp and the MetaModel when we exclude the RDKit descriptors.

\subsection{The limitations of neural networks}

Given the current excitement around deep learning in all its forms, it is understandable that many researchers will use neural networks as a catch-all solution to any problem. However, while ANNs are generally applicable, they are not necessarily generally optimal. 
We find that, on most datasets, a mixed ensemble of ML models can match or outperform ChemProp when both models have access to both external descriptors and ChemProp's latent molecular representation. The best-performing models in the ensemble vary across the datasets (see supplementary information), emphasising the importance of a diverse model set. In many cases, the ANN-based MLP and ResNet models are relatively unimportant to the ensemble predictions, particularly for the classification datasets. 

As both predictive models have access to the same information in both cases, the only explanation for the improved performance of the MetaModel is that the FFN used by ChemProp to turn its latent molecular representations into predictions is underperforming relative to other ML models. Even on datasets where MLP submodels make up a significant fraction of the ensemble (such as QM8 or Lipophilicity), the ensemble still significantly outperforms ChemProp. This suggests that the aggregation process results in a more reliable prediction than relying on a single model or an ensemble of one model class.

There are two potential causes for this. Firstly, ChemProp is likely less able to make use of the external descriptors than other models. This boils down to the well-known issue of predicting tabular datasets with ANNs. These descriptors are unstructured (as opposed to image, text, or graph data, where there are structural relationships between points that ANNs can exploit) and are not always distributed in a way that is easy for FFNs to learn from. This is backed up by the lack of an observed drop in ChemProp performance when the RDKit descriptors are removed, implying it makes minimal use of them. 
Secondly, some target variables are better suited to FFN modelling than others because they align with the model's inductive biases. This is not specific to FFNs; as we see from the analysis of the MetaModel sub-models, different models perform better on different datasets, and no single model class is universally optimal. This is likely the cause of the ChemProp underperforming the MetaModel on the MUV and QM8 datasets, even when external descriptors are not included.

We see a clear distinction in the importance of the MLP and ResNet models between the regression and classification datasets, with these models appearing much less frequently in the final ensembles for classification problems. A potential explanation for this is that classification problems are typically easier for tree-based models to perform well on, as the discontinuities that they naturally produce can mean that they struggle to model smooth regression functions, while classification problems are naturally discontinuous. 

Deep learning models thrive in large-scale predictive tasks with structured data, such as natural language or image recognition. In these cases, no other algorithms can take advantage of the structured data in the same way, and the competition for the state-of-the-art is generally between different ANN architectures. Molecular graphs are similar, in that they are a form of highly structured data, but for many real-world problems the volume of data is much smaller than is typical for language or image datasets. This limits the potential of GNN models to learn all the relevant information, and means that they have to fall back on general-purpose descriptors, which, as we have seen above, they struggle to use as effectively as other ML algorithms.

\subsection{Model choice}

When constructing our MetaModel ensemble, we aimed to include a diverse selection of ML algorithms that would perform well under almost any circumstance. This has worked well across the different datasets, with different models being prioritised to achieve excellent performance throughout. The model aggregation algorithm that we used is relatively simple compared to some other studies\cite{Kwon2019, Vo2023}, and this is intentional to maximise the ability of the MetaModel to generalise. While training a sophisticated model to combine the sub-model predictions may boost performance on large datasets, it also risks introducing an additional source of over-fitting, particularly on small datasets. Because this study primarily focuses on ensuring optimal sub-models are present and prioritised, we have focused more on the sub-model selection than the aggregation process.

Some general trends within the sub-models are evident, with the MLP/ResNet ANN models appearing more frequently in the regression datasets, alongside XGBoost, kernel ridge regression, Gaussian processes and random forests. The QM7 dataset heavily favours tree methods, with 5 gradient boosting models and 2 random forests in the final ensemble. We would expect that the more emphasis the MetaModel places on ANNs, the closer ChemProp and the MetaModel should be in performance, but this is not obviously evident in the data. This may be due to the diversity of the datasets obscuring any such trend.

In the classification datasets, we see that the XGBoost models are heavily favoured, being the largest fraction of models on all datasets except the heavily skewed PCBA, MUV and HIV datasets, which emphasise the naive Bayes and quadratic discriminant analysis models. The fraction of XGBoost models is very high on the BACE dataset, and is likely compromising the overall model performance. Modifying the sub-model pruning method to avoid a single model class from making up over 50\% of the total ensemble would prevent this from occurring and maintain a diverse ensemble, and we will implement this for future studies.

\subsection{Opportunities for future research}
\label{sec:future}

As implemented here, the MetaModel is single-target only, whereas ChemProp can efficiently run multi-target predictions. This can be a potential disadvantage if the dataset has many unrelated targets or an advantage if the dataset contains related targets that can be used to gain more information. We suggest that this is the reason ChemProp performs better relative to the MetaModel on the classification datasets than the regression datasets, as the classification datasets have, on average, a much larger number of targets. Additionally, the Tox21 and ToxCast datasets have many missing values, which reduce the effective size of the dataset for single-target models while still allowing multi-target models to use all rows. Effectively, these datasets are better suited to an imputation model\cite{Irwin20} than the standard QSAR-style MetaModel presented here. 
We anticipate that a modified version of the ensemble approach, capable of multi-target prediction, would perform better on these datasets.

\begin{figure}
    \centering
    \includegraphics[width=0.6\linewidth]{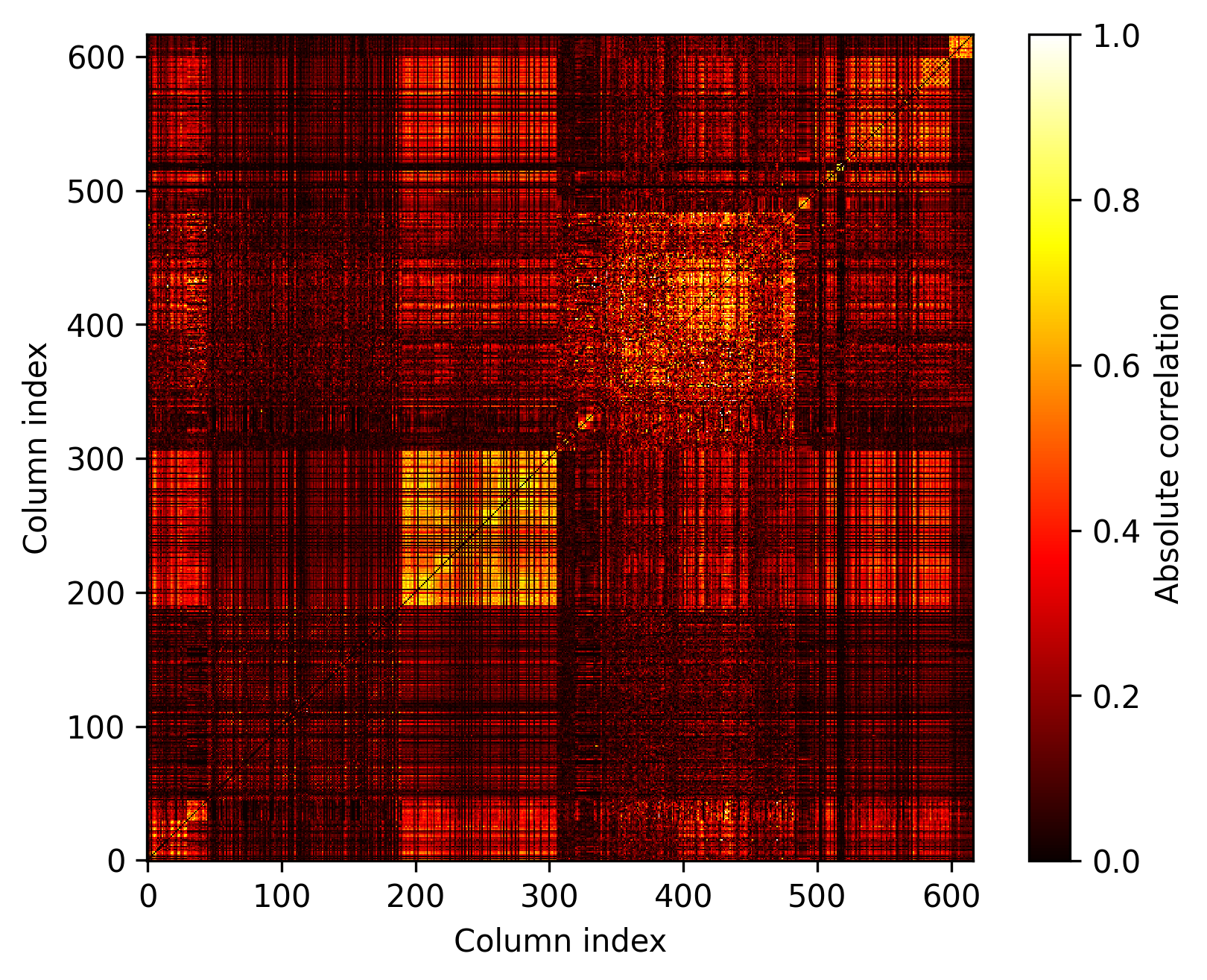}
    \caption{Correlation matrix for the ToxCast dataset. The data is obviously structured, and contains several blocks of columns that correlate with one another. Note that this includes only trivial correlations between targets, and there may be more complex relationships that are not captured by this analysis.}
    \label{fig:toxcast_matrix}
\end{figure}

Taking the ToxCast dataset as an example, the mean number of y values in each target column (and therefore the number of data points available for a single-target model) is 2,235, compared to the full number of 1,378,888 available to a multi-target model. However, not all targets are related, so many of these data points will be useless for predicting any given target. Fig.~\ref{fig:toxcast_matrix} shows the correlation matrix for this dataset, which clearly shows several large groups of correlated variables, ideal for a multi-target model to benefit from. We can derive an approximate value for the number of relevant data points for a given target by multiplying the number of data points in each column by the corresponding correlation coefficient. This gives a mean value of 183,790 relevant data points for each target. Even allowing for redundancy and other effects, this implies that the effective data size for a multi-target model is much larger than for a single target model. Similarly, for Tox21 there are 70,230 total data points, compared to an average of 5,852 in each column. Weighted by the correlations between targets, there is a mean of 18,681 effective data points for a multi-target model.

While it is clear that using GNN features as input descriptors to other models can, in general, give a boost in performance, it is also apparent that there are inefficiencies and trade-offs in our approach. These mainly stem from the relatively crude approach to bolting together the GNN descriptors and the MetaModel ensemble. Our study run without external descriptors demonstrates that the MetaModel is not generally capable of matching the performance of ChemProp on its own latent representations alone, presumably because they are not trained and optimised together (i.e. the latent features are optimised for the FFN prediction head, not for XGBoost, for example).

Similarly, our attempt at devising a hyperparameter optimisation strategy for the combined featurisation plus modelling framework was not successful, and we attribute this to the difficulty of combining optimising one model to maximise the performance of another, separate model. Optimising ChemProp in isolation, ignoring the MetaModel, was also an unsuccessful strategy, resulting in worse MetaModel performance on several datasets. 
This is likely caused by the hyperparameter optimisation process optimising for features of ANNs that do not necessarily transfer over to other model classes, in particular, dropout. In some circumstances, it may be desirable to have the MPNN produce many partially redundant features and then use dropout to prevent the FFN head from overfitting. When these features are used to train an alternative model, without dropout, the large number of features makes the model vulnerable to overfitting and worsens performance. 

Taken together, we are confident that there is still room for further improvement and refinement in the modelling approach outlined in this work. The fundamental benefits of combining task-specific and general-purpose descriptors and using a diverse set of ML models seem robust, but the details of exactly how best to do this are still very uncertain.

\section{Conclusion}

We construct a MetaModel from a weighted ensemble of heterogeneous strong learners and demonstrate a novel featurisation approach using a ChemProp MPNN to derive dataset-specific latent molecular representations. 

Using this, we demonstrate that a diverse set of ML models, trained on a combination of task-specific learned molecular representations and external descriptors, can outperform pure neural network-based approaches on the majority of datasets. We attribute this to neural networks not making optimal use of the external descriptors and to some datasets being challenging to model with a standard feed-forward network. 

We also show that the MetaModel ensemble can make effective use of ChemProp's learned molecular representation, outperforming the same model trained without access to these additional features on most datasets. 

Similarly, we find that excluding the general-purpose descriptor set drops the performance of both ChemProp and the MetaModel on specific datasets, indicating that key information for modelling these datasets is contained in the RDKit descriptors and that ChemProp was unable to derive the same information on the fly from limited data.

We explore two different approaches for hyperparameter optimisation with the combined MPNN featuriser plus MetaModel ensemble, but we find that neither gives a consistent, significant improvement over the baseline models. We suggest that this is because some hyperparameters (such as dropout) that may improve performance of the MPNN in isolation do not result in a better feature set for other models to use.

\section{Data and Software Availability Statement}
All data used in this work is publicly available from the MoleculeNet website (\url{https://moleculenet.org/}) or via the DeepChem API \url{https://github.com/deepchem/deepchem}. The ChemProp library is open-source, and available from \url{https://github.com/chemprop/chemprop}. All sub-models used to construct the MetaModel ensemble are open source, using either SciKit-Learn (\url{https://scikit-learn.org}), SciPy (\url{https://scipy.org/}), PyTorch (\url{https://pytorch.org/}), or XGBoost (\url{https://xgboost.readthedocs.io/en/stable}).

\section{Author Contributions}

MLP: Analysis lead, initial draft, MetaModel project lead.
SM: MetaModel development and benchmarking.
BM: MetaModel benchmarking, manuscript review and editing.
M\"{O}: Project feedback, manuscript review and editing.
HT: MetaModel benchmarking, manuscript review and editing.
CW: MetaModel benchmarking, framework testing.
MDS: Overall project goals, supervision, manuscript review.

\section{Conflict of Interest Statement}
The authors declare no competing financial interests.

\section{Supporting Information}

Details of the sub-models, the datasets, and the bootstrapping method used to calculate significances (PDF).

\begin{acknowledgement}

The authors thank Rae Lawrence for helpful feedback on the manuscript and Ed Champness for useful discussions during model development. This research did not receive any specific grant from funding agencies in the public, commercial, or not-for-profit sectors.


\end{acknowledgement}


\bibliography{bibliography}

\newpage
For Table of Contents Only

\includegraphics[width=3.25in]{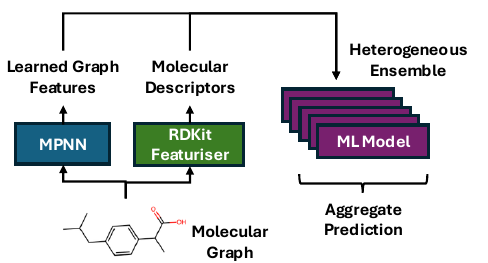}

\end{document}